\documentclass{article}

\usepackage{PRIMEarxiv}

\usepackage[utf8]{inputenc} 
\usepackage[T1]{fontenc}    
\usepackage{hyperref}       
\usepackage{url}            
\usepackage{booktabs}       
\usepackage{amsfonts}       
\usepackage{nicefrac}       
\usepackage{microtype}      
\usepackage{lipsum}
\usepackage{fancyhdr}       
\usepackage{graphicx}       
\graphicspath{{media/}}     
\usepackage{algorithm}
\usepackage{algorithmic}
\usepackage{amsmath}
\usepackage{multirow,makecell}

\title{Knowledge Graph Quality Evaluation under Incomplete Information}

\author{
  Xiaodong Li \\
  Hohai University \\
  Nanjing, China\\
  \texttt{xiaodong.li@hhu.edu.cn} \\
   \And
  Chenxin Zou \\
  Hohai University \\
  Nanjing, China\\
  \texttt{zoucx@hhu.edu.cn} \\
   \And
  Yi Cai \\
  South China University of Technology \\
  Guangzhou, China\\
  \texttt{ycai@scut.edu.cn} \\
   \And
  Yuelong Zhu \\
  Hohai University \\
  Nanjing, China\\
  \texttt{ylzhu@hhu.edu.cn} \\
}

\begin{document}
\maketitle

\begin{abstract}
Knowledge graphs (KGs) have attracted more and more attentions because of their fundamental roles in many tasks. Quality evaluation for KGs is thus crucial and indispensable. Existing methods in this field evaluate KGs by either proposing new quality metrics from different dimensions or measuring performances at KG construction stages. However, there are two major issues with those methods. First, they highly rely on raw data in KGs, which makes KGs' internal information exposed during quality evaluation. Second, they consider more about the quality at data level instead of ability level, where the latter one is more important for downstream applications. To address these issues, we propose a knowledge graph quality evaluation framework under incomplete information (QEII). The quality evaluation task is transformed into an adversarial Q\&A game between two KGs. Winner of the game is thus considered to have better qualities. During the evaluation process, no raw data is exposed, which ensures information protection. Experimental results on four pairs of KGs demonstrate that, compared with baselines, the QEII implements a reasonable quality evaluation at ability level under incomplete information.
\end{abstract}


\section{Introduction}
Knowledge graph (KG) is a graph model-based technology that can describe everything's relations in the world~\cite{2003semantic}. It is not only used for semantic searching but also plays an important role in many other applications, such as intelligent Q\&A~\cite{gao2020knowledge}, personalized recommendation~\cite{xian2019reinforcement} and interpretability for machine learning~\cite{lecue2020role}, etc. Performances of these applications depend on qualities of KGs. Applying a KG with poor quality will result in bad performance. Therefore, quality evaluation for KGs is important and necessary.

Existing methods can be roughly divided into two categories. The first one is to design and propose various quality evaluation metrics from different dimensions and considerations~\cite{zaveri2013user,zaveri2016quality}, where raw data of KGs are exploited as base information for quality evaluation. The second one is to evaluate KGs by measuring performances at construction stages~\cite{li2020real,al2020end}, where KGs those have high performances at different stages through construction are considered to have good qualities. However, there are two major issues with those methods. First, they highly rely on KGs' raw data for quality evaluation, which inevitably results in exposure of KGs' internal information. Second, they concentrate more on KGs' quality at data level but ignore that at ability level, while ability evaluation is much more important for downstream applications. In a word, how to evaluate KGs at ability level while protecting internal information from exposure is a problem worth studying.

In this paper, we propose a knowledge graph quality evaluation framework under incomplete information (QEII) to address those two problems. Inspired by the millionaire problem in secure multiparty computation~\cite{yao1982protocols}, the KG quality evaluation task is transformed into an adversarial Q\&A game between KGs. In the game, each KG makes full use of its knowledge to question the other KG and tries its best to answer questions from the other KG. To protect KGs' raw data from information leak, only minimum and necessary information represented by mutually understandable format is exchanged (i.e., incomplete information). In this assumed scenario, KG quality at ability level is evaluated by their performances in the Q\&A.

The game is divided into two subgames. In the first subgame, a question model ($QM$) and an answer model ($AM$) are trained within each KG, where based on knowledges of each KG the $QM$ generates questions and the $AM$ tries its best to answer the questions. The questions are initially defect subgraphs, which can be directly sampled from KGs without any extra annotation. Difficulties of the questions are quantified by different features, and three question difficulty tuning methods are proposed, i.e., rule-based, Naive Bayes-based, and retrieval-based. According to the methods, questions' difficulties are tuned during training in order to improve the ability of both the $QM$ and the $AM$ in a GAN manner, where the $QM$ adjusts questions' difficulties and generates new batches of questions according to feedbacks from the $AM$, and the $AM$ learns from the questions and improves its answering ability.

In the second subgame, two KGs compete by using their $QM$s to question each other's $AM$. Questions and answers need to be exchanged between KGs. For information protection, a TransE model ($TM$) and two defect subgraph embedding models ($EM$s) are employed. Two KGs train the same $TM$ in an incremental way without exchanging any raw data, where one KG trains the $TM$ and hands it over to the other KG for continual training, vice versa, which makes sure that both KGs encode entities and relations in the questions using the same $TM$. In contrast, they train $EM$s respectively, where one KG uses the other KG's $EM$ to encode its own questions (defect subgraph structure) in order to make sure that the other KG can understand. In this way, only two kinds of information (i.e., questions and answers) and two models (i.e., $TM$ and $EM$) are exchanged during the competition. Q\&A is thus conducted under an incomplete information scenario.

In summary, our contributions are as follows:
\begin{itemize}
	\item We propose a novel knowledge graph quality evaluation framework, which evaluates the quality at ability level through an adversarial Q\&A game under incomplete information, without exposing any internal information of KGs.
	\item In the first subgame, we propose a $QM$ and $AM$ training method in a GAN manner, where questions (defect subgraphs) can be tuned by three difficulty features and three question difficulty adjustment methods.
	\item In the second subgame, we propose a $TM$ and $EM$ training method for information protection, where one $TM$ is incrementally trained and two $EM$s are separately trained, and the $TM$ and $EM$s are further exchanged to facilitate Q\&A under incomplete information.
	\item Experiments are conducted on four pairs of KGs and additional analyses are given. The experimental results demonstrate that, compared with baselines, the proposed QEII is effective, which provides reasonable KG quality evaluation under incomplete information.
\end{itemize}

The remaining sections of this paper are arranged as follows. Section~\ref{sec:two} introduces some work related to our framework. Section~\ref{sec:three} proposes the QEII and introduces two subgames in detail. Section~\ref{sec:four} describes the evaluation experiments and gives some related discussions. Section~\ref{sec:five} concludes the paper and indicates future work.

\section{Related Work}\label{sec:two}
Numerous quality evaluation methods have been proposed for KGs in recent times. These methods can be analyzed and summarized from two dimensions. Horizontal methods evaluate a given KG based on various quality dimensions, while vertical methods assess the quality according to performances at KG construction stages. Section~\ref{sec:two_one} presents a detailed introduction to these methods. Additionally, Sections~\ref{sec:two_two} and~\ref{sec:two_three} highlight other relevant research studies on question generation, question answer, and Generative Adversarial Nets (GAN).

\subsection{Horizontal and Vertical Evaluation}\label{sec:two_one}
Basis of knowledge graphs (KGs) is knowledge, which can be considered as data represented in graphs. Horizontal methods primarily focus on evaluating the quality of knowledge. Zaveri \textit{et al.}~\cite{zaveri2013user,zaveri2016quality} proposed four quality dimensions, namely Accessibility, Intrinsic, Contextual, and Representational, based on common data problems and previous evaluation methods. These four dimensions can be further subdivided to reflect the more fine-grained quality of data. Chen \textit{et al.}~\cite{chen2019practical} extended Diversity and Robust dimensions under Contextual and Representational, respectively, by analyzing common applications of KGs. Evaluation of data always considers multiple dimensions together based on demands. Xue \textit{et al.}~\cite{xue2022knowledge} emphasized on Intrinsic and Timeliness under Contextual and surveyed previous quality management methods. Debattista \textit{et al.}~\cite{debattista2016luzzu} paid attention to five dimensions under Contextual and Intrinsic, and Färber \textit{et al.}~\cite{farber2018knowledge,farber2018linked} focused on eleven dimensions. A comprehensive evaluation was obtained by integrating metrics in different designed manners. Semantic accuracy is a commonly evaluated fine-grained dimension under Intrinsic. Gao \textit{et al.}~\cite{gao2019efficient} clustered triples in KGs based on subjects and sampled some triples from each cluster for evaluation, which helped reduce manual annotation costs. Kontokostas \textit{et al.}~\cite{kontokostas2014test} evaluated semantic accuracy by constructing SPARQL queries regarding six possible accuracy problems and calculating an error rate in returned cases.

Vertical methods are utilized to evaluate KGs by measuring performances at construction stages. Knowledge extraction is a crucial stage that mainly focuses on the correctness and completeness of KGs~\cite{fensel2020build,paulheim2017knowledge}. The correctness is typically evaluated through precision, recall, and F1~\cite{li2020real,liu2020preliminary,jiang2019role}. These metrics require annotated data for their calculation, and in the absence of it, triples are manually reviewed to determine whether to be correct~\cite{martinez2018openie,yu2017knowledge}. The limited number of triples cannot represent the vast amount of knowledge. The completeness is often assessed by some statistics, such as the numbers of triples, entities, and relations present~\cite{chaves2019parameters,yoo2020automating}. Another significant construction stage is KG application. It applies KGs to downstream applications, whose performances show the quality of KGs~\cite{al2020end,clancy2019knowledge}. These methods can also be considered as evaluating the quality of given KGs, which is consistent with the definition of horizontal evaluation.

\subsection{Question Generation and Question Answer}\label{sec:two_two}
Question generation involves generating questions in natural language from given answers. Template-based methods ensure that the questions generated are easily readable. Duan \textit{et al.}~\cite{duan2017question} collected questions from webpages and removed frequent n-grams to create a template. Appropriate templates and n-grams were then chosen via retrieval-based CNN or generation-based RNN to generate questions. Zeng \textit{et al.}~\cite{zeng2020exploiting} similarly used predefined mappings from triple predicates to templates, which would generate a question when subjects were filled in. However, these methods often require complex natural language processing and rule design, which are costly. To address this issue, encoder-decoder models have been developed for question generation in a seq2seq manner. Encoder is always designed based on LSTM~\cite{du2017learning}. To generate diverse and fluent questions, the encoder needs to capture more semantics from some external information, such as descriptions about entities in Wikidata~\cite{bi2020knowledge} and neighbors in KGs~\cite{indurthi2017generating}. Decoder is used to decode the output of the encoder in different modes and generate questions, with copy mode and generate mode~\cite{wang2018neural} being the most commonly used. Generated questions may have different difficulties although their answers are the same. To define the question difficulty, Kumar \textit{et al.}~\cite{kumar2019difficulty} used confidence of entity links and selectivity of entities, whereas Seyler \textit{et al.}~\cite{seyler2015generating} based their definition on popularity, selectivity, and coherence.

Question answer involves understanding the given question and providing an appropriate answer. In the past, researchers such as Lin \textit{et al.}~\cite{lin2003question} relied on Google to retrieve relevant documents and extract answers. KGs have significantly improved machines' ability to understand questions. To obtain triples that are similar to questions, Bao \textit{et al.}~\cite{bao2014knowledge} calculated a correlation score to rank triples based on linear features captured from candidate triples, knowledge base, and questions. Yin \textit{et al.}~\cite{yin2016neural} leveraged a bilinear model or CNN-based matching model to calculate a correlation score between questions and candidate triples. External knowledge, such as relevant documents from search engines~\cite{xiong2019improving}, is also commonly incorporated into the answering process. In addition, questions can be reformulated into machine-readable formats. Cui \textit{et al.}~\cite{cui2017kbqa} replaced entities with their concepts, effectively converting questions into templates. Sorokin \textit{et al.}~\cite{sorokin2018modeling} proposed graph-based representations for questions. They identified the entities in questions and linked them to an initialized question node with relevant relations in KGs. Vakulenko \textit{et al.}~\cite{vakulenko2019message} suggested using subgraphs that contained entities, predicates, and their types from KGs as question representations. Answers were then obtained through message transfers between nodes in the subgraphs.

\subsection{GAN}\label{sec:two_three}
GAN~\cite{goodfellow2014generative} is a type of adversarial model used to generate distributions of real data. It consists of a generator and a discriminator, where the generator creates images based on input noise while the discriminator identifies whether a given image is real or generated. Both of them are trained simultaneously and adversarially. However, GAN has difficulties modeling discrete data, making text generation problematic. To address this issue, Zhang \textit{et al.}~\cite{zhang2017adversarial} used CNN as the discriminator and LSTM as the generator. They proposed to capture features of true and false text and calculated Maximum Mean Discrepancy between them as an objective function. Kusner \textit{et al.}~\cite{kusner2016gans} transformed discrete variables into continuous ones through Gumbel and softmax. Yu \textit{et al.}~\cite{yu2017seqgan} and Guo \textit{et al.}~\cite{guo2018long} optimized the generator using reinforcement learning, where states were the generated words and action was to generate the next word. Rewards were given by the discriminator.

Existing methods heavily rely on raw data within KGs in order to design evaluation metrics. However, they often result in information exposure and the metrics designed at data level are not sufficient for evaluating KGs' abilities. Differently, we propose an adversarial game between two KGs to evaluate their qualities at ability level through Q\&A. In the game, each KG trains a $QM$ and an $AM$ in a GAN manner. Instead of using natural language questions, we consider defect subgraphs as a new question representation. These subgraphs can be directly sampled from KGs, eliminating the need for annotated data. During training, the $QM$ generates new questions based on feedback from the $AM$, which is different from the above question generation task; the $AM$ learns to answer questions from the $QM$, similar to the question answer task. KGs then exchange questions generated by their respective $QM$s and use their $AM$s to answer them. Both models are used for quality evaluation. It is different from the typical GAN where the generator is trained for image generation and the discriminator acts as an assistant for training.

\section{QEII}\label{sec:three}
\begin{figure}[!t]
	\centering
	\includegraphics[width=0.7\linewidth]{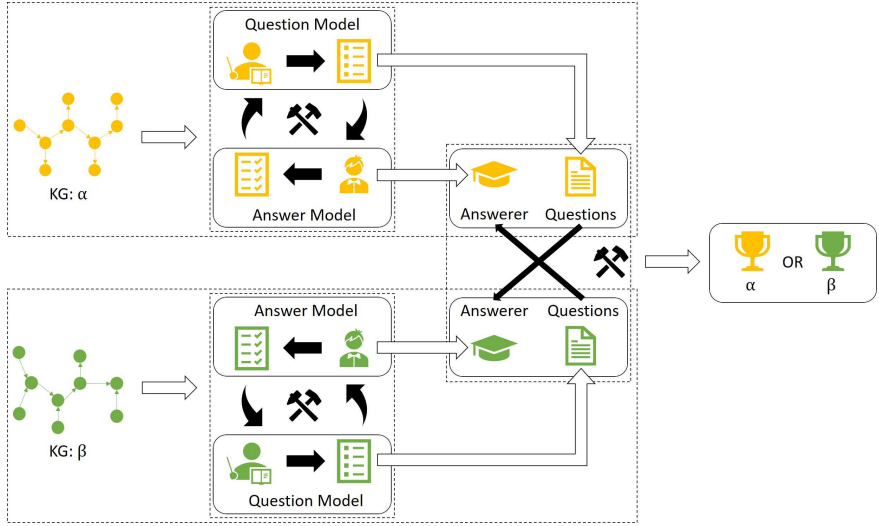}
	\caption{Workflow of the QEII.}
	\label{fig:QEII}
\end{figure}

In the QEII, a $QM$ and an $AM$ are trained within each KG for Q\&A, and the quality is evaluated in terms of scores obtained by answering the exchanged questions. The whole evaluation process is shown in Figure \ref{fig:QEII}, which can be considered an adversarial game that consists of two subgames. The first subgame is between $QM$ and $AM$ of a KG, and the second one is between KGs.

\subsection{Task Definition}
Two KGs to be evaluated are $\alpha$ and $\beta$. In the first subgame, the $QM$ samples defect subgraphs $sg$ as questions and generates different question sets $Q_t$ $(t=0,1,2\dots)$ according to difficulty features $L$. A defect subgraph refers to the subgraph sampled from KGs, where attribute or entity information of a node is removed and it is converted to a blank node. A question in $Q_t$ is $qa=\{q,A\}$, where $q$ is a description of the question and $A$ is a candidate answer set. There are two types of questions in $Q_t$: $qa$ is a judgment question if $|A|=1$, i.e., $A=\{a_c\}$ or $A=\{a_w\}$, where $a_c$ is the correct answer and $a_w$ is the wrong answer; $qa$ is a choice question if $|A|>1$, i.e., $A=\{a_c,a_{w1},\dots,a_{wn_w}\}$. The $AM$ answers the questions in $Q_t$. While answering judgment questions, the $AM$ judges whether the candidate answer is correct or not. While answering choice questions, the $AM$ selects an answer from $A$.

In the second subgame, $\alpha$ and $\beta$ train the same $TM$ in an incremental way but train their own $EM^{\alpha}$ and $EM^{\beta}$ separately. Both KGs exchange the $TM$ during training to embed entities and relations. Their $EM$s are also exchanged to further represent question descriptions as vectors after training. While KGs Q\&A each other, $\alpha$ uses $TM$ and $EM^{\beta}$ to encode $Q^{\alpha}$ for $\beta$. $\beta$ answers these questions using $AM^{\beta}$ and returns answers $A_{\#}^{\beta}$ to $\alpha$. $\alpha$ reviews $A_{\#}^{\beta}$ according to correct answers and gives a score $S^{\beta}$. Similarly, $\beta$ gives a score $S^{\alpha}$ after reviewing the answers from $\alpha$. The qualities of $\alpha$ and $\beta$ are evaluated by comparing $S^{\alpha}$ and $S^{\beta}$.

\subsection{Subgame between QM and AM}\label{sec:three_two}
\begin{figure}[!t]
	\centering
	\includegraphics[width=0.7\linewidth]{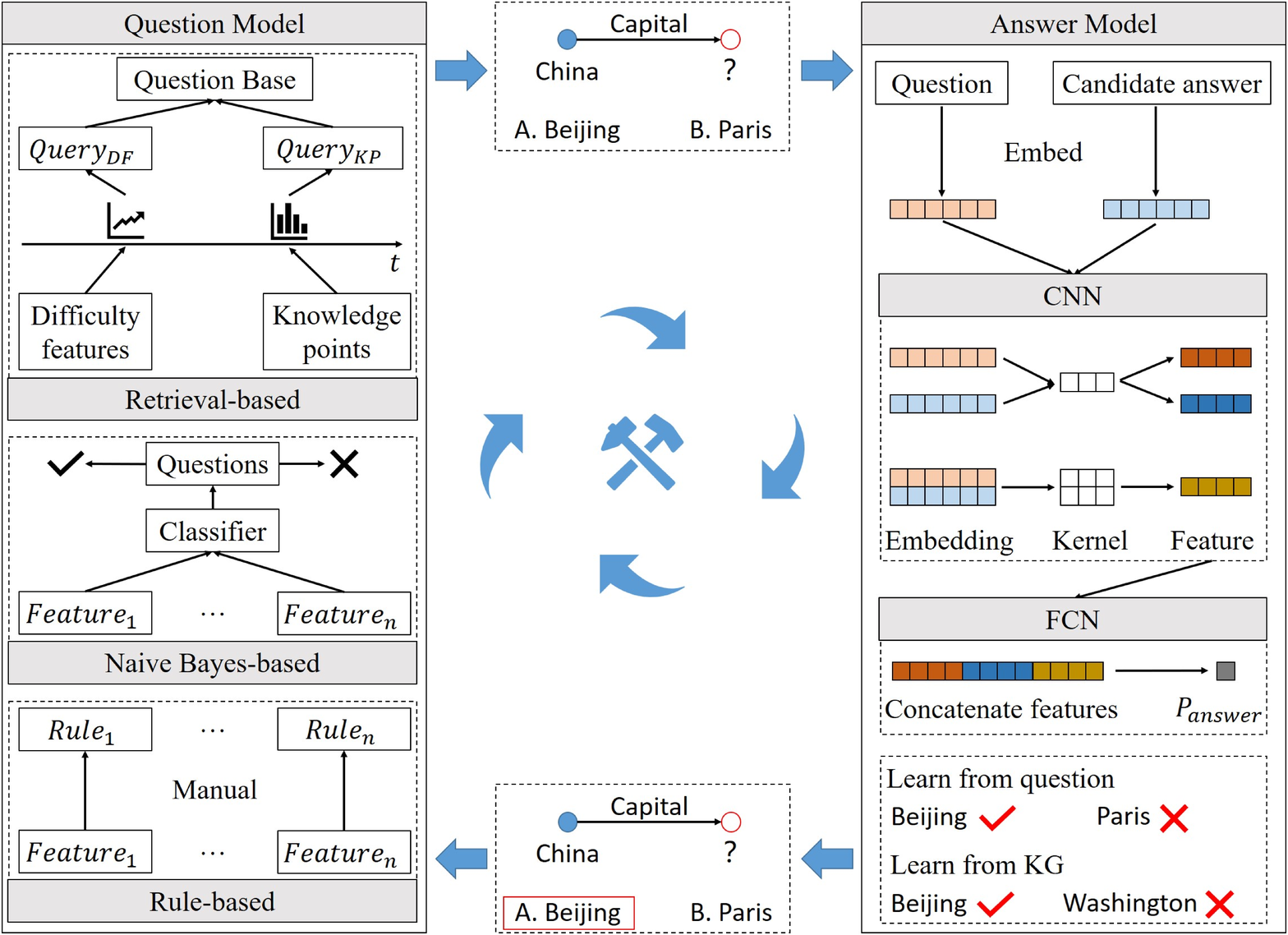}
	\caption{Workflow of the first subgame.}
	\label{fig:adversary1}
\end{figure}

Workflow of the first subgame is shown in Figure \ref{fig:adversary1}. Each KG makes full use of its knowledge to train a $QM$ and an $AM$ in a GAN manner. Specifically, there are four problems to be solved: 1) how to generate questions, 2) how to answer questions, and 3) how to train and optimize $QM$ and $AM$ simultaneously.

\subsubsection{Question Generation}
During the training of models, the $QM$ generates questions based on feedback received from the $AM$ and the $AM$ then responds to the questions generated by the $QM$. Similar to GAN, the $QM$ and the $AM$ are opponents, and there are many rounds of competitions between them until both models are of good performance. In the initial round (i.e., the $0$-th round), the $QM$ generates a set of questions, denoted as $Q_0$, in which each question is represented as a defect subgraph. Different from natural languages, it can be directly sampled from KGs without any annotated data. Inspired by examination papers, we adopt two types of questions: judgment questions and choice questions. The primary difference between these two types is the number of candidate answers. Questions with a single candidate answer are judgment questions, while those with multiple candidate answers are choice questions.

While generating $Q_0$, sets $O$ and $C_{sg}$ are defined to sample the number of candidate answers in questions and nodes in subgraphs respectively in the first step. A subgraph is then sampled through a random walk starting from a randomly selected node in KGs. The attribute or entity information of a node in the subgraph is removed and it is converted to a blank node, generating defect subgraph that serves as a question description. The removed information from the blank node represents a correct answer, and several wrong answers are sampled from KGs to generate a set of candidate answers.

It is important to consider that in some cases, randomly selected candidates may actually be correct answers due to the presence of 1-N relations in KGs. There are many triples $(s,p,o_i)$ if $p$ is a 1-N relation, where multiple objects are associated with the same subject-predicate pair. All $o$s cannot be sampled as wrong answers if one of them is correct.

\subsubsection{Question Answer}
While answering questions in an embedding space, $\vec{s}+\vec{p}=\vec{o}$ in the triple $\left(\vec{s},\vec{p},\vec{o}\right)$ is often employed to search for the correct answer. If a question contains an entity $s$ and a relation $p$, an approximate embedding $\vec{o}$ of the correct answer can be calculated as $\vec{s}+\vec{p}$. Similarly, candidate answer embedding $\mathbf{a}^c$ and question embedding $\mathbf{FSG}$ are used to judge whether the candidate is correct. Convolutional Neural Network (CNN) is leveraged to capture features of $\mathbf{a}^c$ and $\mathbf{FSG}$, and Fully Convolutional Network (FCN) is used to calculate a probability that the candidate answer is correct. Specifically, $\mathbf{FSG}$ and $\mathbf{a}^c$ are concatenated at first. Then, convolution kernels are used to capture features,
\begin{equation}
\mathbf{FA}=LeakyReLU\left(\sum_{x\in fm}xk+b\right),fm\in\mathbf{QA},k\in K,
\end{equation}
\begin{equation}
\mathbf{QA}=\left(\begin{array}{c}
\mathbf{FSG}\\\mathbf{a}^c
\end{array}\right),
\end{equation}
where $\mathbf{FA}$ is the feature captured by the convolution kernel $K$ and $fm$ is the feature map from $\mathbf{QA}$. There are two different convolution kernels: one is to capture local features $\mathbf{FA}_1$ of defect subgraphs and candidate answers, and the other is used for global features $\mathbf{FA}_2$ of their combination. Finally, $\mathbf{FA}_1$ and $\mathbf{FA}_2$ are concatenated to calculate a probability $P^c$ through FCN,
\begin{equation}
P^c=Sigmoid\left(W_0\left[\mathbf{FA}_1;\mathbf{FA}_2\right]+b_0\right),
\end{equation}
where $W_0$ is a weight matrix and $b_0$ is a bias. While answering a judgment question, the probability $P_{JQ}^c$ is calculated and the only candidate answer is correct if $P_{JQ}^c>0.5$. While answering a choice question, all candidate answers are traversed to calculate the probability $P_{CQ_i}^c$. The correct answer is the index corresponding to the maximum probability $\mathop{\arg\max}\limits_i P_{CQ_i}^c$.

\subsubsection{Adversarial Training between QM and AM}
In the $t$-th round of competitions between $QM$ and $AM$ during training, the $QM$ generates $Q_t$ and the $AM$ answers $Q_t$, resulting in $\{Q_t^+,Q_t^-\}$ where $Q_t^+$ is the question set answered correctly and $Q_t^-$ is the set answered wrongly. The $QM$ is then to generate $Q_{t+1}$ according to $\{Q_t^+,Q_t^-\}$. Question difficulty is introduced as the guide for question generation. Suppose that the question difficulty depends on multiple features $\{l_1,l_2, \dots,l_{n_L}\}$, then the difficulty of a question is characterized by $L=\left(\mu_1,\mu_2,\dots,\mu_{n_L}\right)$, where $\mu_i$ is the value of $i$-th feature $l_i$ and $n_L$ is the total number of features. The $QM$ controls answering accuracy of the $AM$ by adjusting question difficulty. The $AM$ correctly answers as many questions from the $QM$ as possible and improves the accuracy to the maximum. The training stops while the accuracy is within a range $\eta$.

For the $QM$, we propose three question difficulty tuning methods: rule-based, Naive Bayes-based, and retrieval-based. The rule-based method involves analyzing relationships between question difficulty and each feature $l_i$, determining how the difficulty changes as feature values increase or decrease. Then, two tuning rules $r_i^+$ and $r_i^-$ are developed based on $l_i$, with $r_i^+$ increasing feature values and $r_i^-$ decreasing values. There are different rules for different difficulty features. Different applying orders of rules are defined to improve and reduce question difficulty respectively. Finally, the accuracy $acc$ of the $AM$ is calculated and compared with $\eta$,
\begin{equation}
acc = \frac{|Q_t^+|}{|Q_t|}.
\end{equation}
If $acc>\eta$, the rules are used to improve the difficulty of questions in $Q_t^+$ and generate a new question set $Q_t^{'+}$. The union of $Q_t^-$ and $Q_t^{'+}$ is considered as $Q_{t+1}$. If $acc<\eta$, the rules are used to reduce the difficulty of questions in $Q_t^-$ and generate $Q_t^{'-}$ to obtain the union $Q_{t+1}=Q_t^+\cup Q_t^{'-}$.

The rule-based method described above offers a stable means of tuning question difficulty without introducing any elements of randomness, but it requires manual design for ${r_i^+}$ and ${r_i^-}$, which brings lots of costs. Therefore, a more automated method is further explored to generate questions based on their difficulty. Specifically, in the $t$-th round of competitions, the results $\{Q_t^+,Q_t^-\}$ of the $AM$ are used as annotated data to train a classifier, and the trained classifier is able to predict whether a question can be answered correctly or not. Different from tuning questions directly, the classifier is used to select questions rather than change them.

As a classifier, Naive Bayes model calculates feature distributions based on probability theory. It can be used to model difficulty features for question prediction. In the Naive Bayes-based method, more questions are sampled from KGs to extend $Q_0$ as a question base $QB$ at first. All questions in $Q_t$ $(t\geq 1)$ are sourced from $QB$. Then, the questions in $Q_t^+$ and $Q_t^-$ are labeled as $R^+$ and $R^-$ respectively. Values of the $i$-th difficulty feature for all questions in $QB$ are $\{\mu_1,\mu_2,\dots,\mu_{n_{l_i}}\}$. We suppose that difficulty features are mutually independent. The prior probability that a question is answered correctly or wrongly is,
\begin{equation}
P\left(Y=R^*\right)=\frac{\sum_{k=1}^{N_{Q_t}}I\left(y_k=R^*\right)}{N_{Q_t}},
\end{equation}
and distributions of the difficulty features in each category of labeled questions are calculated as,
\begin{equation}
P\left(l_i=\mu_j|Y=R^*\right)=\frac{\sum_{k=1}^{N_{Q_t}}I\left(l_i^k=\mu_j,y_k=R^*\right)}{\sum_{k=1}^{N_{Q_t}}I\left(y_k=R^*\right)},
\end{equation}
where, $Y$ is the label of a question, $N_{Q_t}$ is the number of questions in $Q_t$, and $I\left(\cdot\right)$ is an indicator function. The labels of questions in $QB$ except $Q_t$ are predicted in terms of the above probabilities,
\begin{equation}
y=\mathop{\arg\max}\limits_{R^*}P\left(Y=R^*\right)\prod_{i=1}^{n_L}{P\left(l_i=\mu_j|Y=R^*\right)}.
\end{equation}
Finally, a ratio is established to select questions categorized as $R^+$ and $R^-$ based on $\eta$, resulting in a revised question set $Q_{t+1}$ derived from $QB$. $Q_{t+1}$ contains the same number of questions as $Q_t$, but both are fewer in quantity than those in $QB$, i.e., $N_{Q_{t+1}}=N_{Q_t}<N_{QB}$.

In comparison to the rule-based method, the Naive Bayes-based method provides automatic generation of a new question set based on feedback from the $AM$, resulting in a significant reduction of costs for rule design. However, two issues need to be addressed. Firstly, it is necessary to distinguish discrete and continuous features. While distributions of the discrete features can be directly derived from their values, those of continuous features require fitting by a standard function whose parameters are estimated based on feature values. Different approaches are adopted to calculate these distributions. Secondly, incorporating more information is challenging. Possible correlations among different information undermine the independent supposition and, consequently, reduce the classification performance of the model.

In the retrieval-based method, the question generation task is modeled as an information retrieval process. Both implicit and explicit information are utilized to query questions. The implicit information refers to question difficulties and the explicit information refers to knowledge points, i.e., nodes in defect subgraphs. A question $qa$ is represented as $qa_l=\left(\mu_1,\mu_2,\dots,\mu_{n_L}\right)$ based on question difficulties or $qa_v=\left(\phi_1,\phi_2,\dots,\phi_{n_V}\right)$ based on knowledge points, where $\mu_i$ is the value of $l_i$ and $\phi_i\in\{0,1\}$ indicates the presence or absence of a knowledge point $v_i$. First, question difficulty queries $U_l=\{u_l^+,u_l^-\}$ and knowledge point queries $U_v=\{u_v^+,u_v^-\}$ are constructed, where $u^+$ or $u^-$ are used to retrieve the questions that can be correctly or wrongly answered. It is the same to construct $u^+$ and $u^-$, so only $u^+$ is introduced below.

While constructing $u_l^+=(\mu_1^+,\mu_2^+,\dots,\mu_{n_L}^+)$, Gaussian function is used to fit distributions of $l_i$,
\begin{equation}
\Gamma\left(l_i;\theta_1,\theta_2\right)=\frac{1}{\sqrt{2\pi}\theta_2}e^{-\frac{\left(l_i-\theta_1\right)^2}{2{\theta_2}^2}},
\end{equation}
\begin{equation}
\theta_1,\theta_2=\mathop{\arg\max}\limits_{\theta_1,\theta_2}\prod_{t=1}^{T}\prod_{qa\in Q_t^+}\Gamma\left(\mu_i\right),
\end{equation}
where $\theta_1$ and $\theta_2$ are parameters calculated through maximum likelihood estimation. $\mu_i^+=\theta_1$ is the value of $l_i$ in the questions that the $AM$ is most likely to answer correctly. All question difficulty features are idealized as continuous features whose distributions are calculated following the above steps. This idealization may bring errors in calculated feature distributions, which mistakes label predictions in the Naive Bayes model. However, differences caused by these errors are consistent across all questions, which does not alter question ranks in the retrieval results returned by queries. While constructing $u_v^+=\left(\phi_1^+,\phi_2^+,\dots,\phi_{n_V}^+\right)$, the frequency $f_i$ of $v_i$ is calculated by aggregating data on all questions answered correctly or wrongly, and $\phi_i$ is obtained by comparing $f_i^+$ and $f_i^-$,
\begin{equation}
f_i^+=\frac{\sum_{t=1}^{T}\sum_{qa\in Q_t^+}I\left(v_i=1\right)}{\sum_{t=1}^{T}N_{Q_t^+}},
\end{equation}
\begin{eqnarray}
\phi_i^+ = \left\{\begin{array}{ll}
1, & \text{if}\;f_i^+ > f_i^-, \\
0, & \text{otherwise}.
\end{array}\right.
\end{eqnarray}
Then, according to different characteristics of the queries, vector model is used to measure the correlations between question difficulty queries and questions, which can be quantified through cosine similarity,
\begin{equation}
sim_l\left(u_l,qa_l\right)=\frac{\sum_{i=1}^{n_L}\mu_{i,u}\times\mu_{i,qa}}{\sqrt{\sum_{i=1}^{n_L}\mu_{i,u}^2}\times\sqrt{\sum_{i=1}^{n_L}\mu_{i,qa}^2}}.
\end{equation}
Probability model is used for similarity between knowledge point queries and questions,
\begin{equation}
\begin{aligned}
sim_v(u_v,qa_v)&=&&P(\mathcal{R}|u_v,qa_v)\\
&=&&\frac{P(qa_v|\mathcal{R},u_v)\cdot P(\mathcal{R},u_v)}{P(u_v,qa_v)}\\
&=&&\frac{P(qa_v|\mathcal{R},u_v)\cdot P(\mathcal{R}|u_v)}{P(qa_v|u_v)}\\
&=&&P(\mathcal{R}|u_v)\cdot\frac{\prod_{i=1}^{n_V}P(v_i|\mathcal{R},u_v)P(\bar{v_i}|\mathcal{R},u_v)}{\prod_{i=1}^{n_V}P(v_i|u_v)P(\bar{v_i}|u_v)}\\
&=&&\frac{\mathcal{R}}{N}\cdot\prod_{i=1}^{n_V}\frac{\frac{r_i}{\mathcal{R}}(1-\frac{r_i}{\mathcal{R}})}{\frac{n_i}{\mathcal{R}}(1-\frac{n_i}{\mathcal{R}})}\\
&=&&\frac{\mathcal{R}^{1-n_V}}{N^{1-n_V}}\cdot\prod_{i=1}^{n_V}\frac{r_i(\mathcal{R}-r_i)}{n_i(N-n_i)},\\
\end{aligned}
\end{equation}
where,
\begin{equation}
P(v_i|\mathcal{R},u_v)=\frac{r_i}{\mathcal{R}},
\end{equation}
\begin{equation}
P(v_i|u_v)=\frac{n_i}{N},
\end{equation}
\begin{equation}
P(\mathcal{R}|u_v)=\frac{\mathcal{R}}{N},
\end{equation}
$\mathcal{R}$ is the ideal question set correlated to $u_v$ and $P(\cdot)$ is the probability whose meaning is shown in Table \ref{tab:probability}. Knowledge points are conditionally independent of each other. Quantitative relationships in the above equations are detailed in Table \ref{tab:equation}.
\begin{table}[!t]
	\caption{Meanings of probability equations.}
	\label{tab:probability}
	\centering
	\begin{tabular}{lc}
		\toprule
		Equations & Probabilities \\
		\midrule
		$P(\mathcal{R}|u_v,qa_v)$ & $qa$ is correlated to $u_v$ \\
		$P(qa_v|\mathcal{R},u_v)$ & \makecell[c]{a question randomly sampled from $\mathcal{R}$ \\ is represented as $qa_v$} \\
		$P(qa_v|u_v)$ & \makecell[c]{a question randomly sampled from $QB$ \\ is represented as $qa_v$} \\
		$P(\mathcal{R}|u_v)$ & \makecell[c]{a question randomly sampled from $QB$ \\ is correlated to $u_v$} \\
		$P(v_i|\mathcal{R},u_v)$ & a question randomly sampled from $\mathcal{R}$ contains $v_i$ \\
		$P(\bar{v_i}|\mathcal{R},u_v)$ & \makecell[c]{a question randomly sampled from $\mathcal{R}$ \\ does not contain $v_i$} \\
		$P(v_i|u_v)$ & a question randomly sampled from $QB$ contains $v_i$ \\
		$P(\bar{v_i}|u_v)$ & \makecell[c]{a question randomly sampled from $QB$ \\ does not contain $v_i$} \\
		\bottomrule
	\end{tabular}
\end{table}
\begin{table}[!t]
	\caption{Quantitative relationships in equations.}
	\label{tab:equation}
	\centering
	\begin{tabular}{lccc}
		\toprule
		& Related questions & Unrelated questions & Total \\
		\midrule
		Contain $v_i$ & $r_i$ & $n_i-r_i$ & $n_i$ \\
		Not contain $v_i$ & $\mathcal{R}-r_i$ & $(N-n_i)-(\mathcal{R}-r_i)$ & $N-n_i$ \\
		Total & $\mathcal{R}$ & $N-\mathcal{R}$ & $N$ \\
		\bottomrule
	\end{tabular}
\end{table}
Finally, a compromise parameter is introduced to get a comprehensive similarity,
\begin{equation}
sim\left(u,q\right)=\gamma\cdot sim_l\left(u_l,qa_l\right)+\left(1-\gamma\right)\cdot sim_v(u_v,qa_v).
\end{equation}
There are also multiple parameters if more kinds of information are considered for question query. The correlations between this information can be weakened or strengthened by adjusting compromise parameters. All questions in $QB$ are sorted in a descending rank based on $sim\left(u^+,q\right)$ and $sim\left(u^-,q\right)$, and top-$k$ questions are chosen as $Q_{t+1}$ according to $\eta$.

For the $AM$, we propose two learning methods. One is to learn solely from questions without any other knowledge, and the other is to learn more relevant knowledge from KGs. The $AM$ is primarily based on the second method and assisted by the first one. Specifically, the $AM$ pays attention to the questions in $Q_t$ at first and learns which candidate answers are correct and which are wrong. Then, more wrong answers are sampled from KGs, and the $AM$ further learns the difference between these wrong answers and the correct answer, which helps improve the accuracy to answer similar questions. By combining these two learning methods, the $AM$ can effectively utilize the information from $QM$ to enhance its performance.

After undergoing several rounds of competitions, the $AM$'s ability is improved but the accuracy may be greater or less than $\eta$ because difficulty of each question set is different. There is a balance between the answering ability of the $AM$ and the difficulty of questions generated by the $QM$ while the accuracy is within the target range. Any attempts to further enhance the ability or the difficulty may result in decreased performance for the $QM$ or the $AM$, ultimately leading to failure in the game.

\subsection{Subgame between KGs}\label{sec:three_three}
\begin{figure}[!t]
	\centering
	\includegraphics[width=0.7\linewidth]{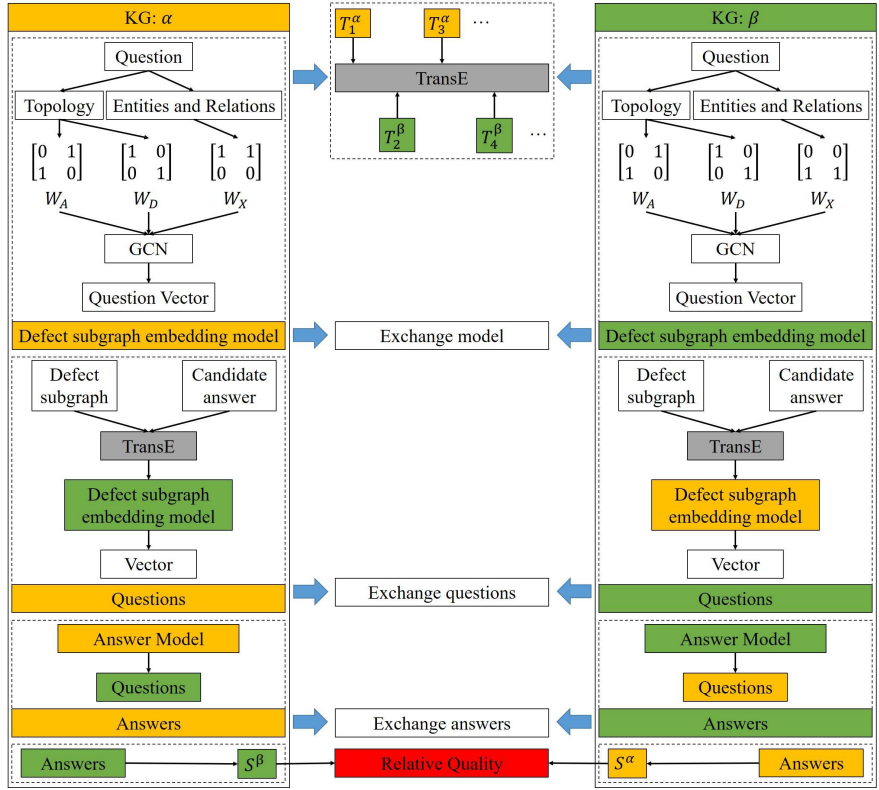}
	\caption{Workflow of the second subgame.}
	\label{fig:adversary2}
\end{figure}

Workflow of the second subgame is shown in Figure \ref{fig:adversary2}. KGs exchange questions generated by their $QM$s and use the $AM$s to answer the exchanged questions. Quality is evaluated according to scores. There are four problems to be solved: 1) what information is necessary to be exchanged, 2) how to represent this information under incomplete information, 3) how to construct this representation, and 4) how to ensure the representations can be understood.

In contrast to the first subgame, there is necessary information needed to exchange between KGs. This information includes questions and answers. The answers are some labels that do not involve internal information. The questions are represented as defect subgraphs. Relevant information about the KG who generates these questions will be exposed while exchanging defect subgraphs directly. To protect the information conveyed by these defect subgraphs, they are further represented as vectors using nonlinear encoding. This ensures that KGs are unable to decode the exchanged question vectors to access each other's internal information.

A two-step embedding method is proposed to learn vector representations. In the first step, two KGs to be evaluated are used to train a $TM$. TransE, a commonly used KG embedding model, is employed to map entities and relations to a vector space in terms of $\vec{s}+\vec{p}=\vec{o}$. First, dimension of the vector space is defined as $n_{VS}$, and a vector for each entity and relation is initialized. Then, the head or tail entity of triples $Tr$ in KGs is replaced according to a random probability following the uniform distribution to form negative samples $Tr^-$,
\begin{eqnarray}
(s,p,o)\to \left\{\begin{array}{ll}
(s^-,p,o), & P_{Tr}>0.5, \\
(s,p,o^-), & \text{otherwise},
\end{array}\right.
\end{eqnarray}
where, $\left(s,p,o\right)$ is a positive sample, $\left(s^-,p,o\right)$ and $\left(s,p,o^-\right)$ are generated negative samples, $s^-$ and $o^-$ are randomly selected from the entity set, and $P_{Tr}$ is a random number generated on (0,1] following the uniform distribution. Finally, the vectors of entities and relations are obtained by minimizing the loss function,
\begin{equation}
loss_{TransE}=\sum_{(s,p,o)\in Tr}\sum_{(s^-,p^-,o^-)\in Tr^-}\left[\xi+d(\vec{s}+\vec{p},\vec{o})-d(\vec{s^-}+\vec{p^-},\vec{o^-})\right]_+,
\end{equation}
where, $\xi$ is a hyper-parameter, $\left[x\right]_+=\max{\left(0,x\right)}$, and $d\left(\vec{x},\vec{y}\right)=\left(\vec{x}-\vec{y}\right)^2$. To protect internal information from exposure, the $TM$ is trained in an incremental way, where one KG trains it and hands it over to the other KG for continual training, and vice versa. KGs are only accessible to their own triples but not those of the other one.

In the second step, an $EM$ based on Graph Convolutional Network (GCN) is proposed to further calculate vector representations of questions. In a defect subgraph, there is a blank node whose attribute or entity information has been removed but its position information is still preserved. Adjacency matrices of the subgraphs remain the same after the removal of node information. In the feature matrix, representation of the blank node is $\textbf{0}$. First, the feature matrix $W_X$, adjacency matrix $W_A$, and degree matrix $W_D$ of a defect subgraph are calculated according to the trained $TM$. Then, GCN captures local and global topology information to generate node representations,
\begin{equation}
\mathbf{H}_{GCN}=ReLU(\widetilde{W_A}W_XW_1),
\end{equation}
\begin{equation}
\mathbf{Z}^{sg}=\widetilde{W_A}\mathbf{H}_{GCN}W_2=\left(\begin{array}{ccc}
z_{0,0}^{sg}&\cdots&z_{0,q}^{sg}\\\vdots&\ddots&\vdots\\z_{N_{sg},0}^{sg}&\cdots&z_{N_{sg},q}^{sg}
\end{array}\right),
\end{equation}
where, $\widetilde{W_A}={W_D}^{-\frac{1}{2}}W_A{W_D}^{-\frac{1}{2}}$, $W_1$ and $W_2$ are weight matrices, and $N_{sg}$ is the number of nodes in the defect subgraph $sg$. Finally, average pooling is used for $\mathbf{Z}^{sg}$ to generate the defect subgraph representation $\mathbf{FSG}$,
\begin{equation}
\mathbf{FSG}=\frac{1}{N_{sg}}\sum_{i=0}^{N_{sg}}\mathbf{z}_{i,\cdot}^{sg}.
\end{equation}
By applying average pooling, the dimensionality of defect subgraph representations is effectively reduced while preserving all node features, thereby enhancing fault tolerance. Each KG trains an $EM$ using its own knowledge. To facilitate mutual comprehension of question vectors, KGs exchange their $EM$s and calculate representations of the questions to be exchanged through the same $TM$ and the $EM$ of the other one. The $AM$s are able to answer the questions in such representations.

Algorithm \ref{alg:QEII} illustrates the entire adversarial game, which takes as input two KGs to be evaluated. First, a $TM$ is trained in an incremental way. Two question sets are then sampled from each KG, where one set is employed to jointly train $EM$ and $AM$, while the other set is used for adversarial training of $QM$ and $AM$. Finally, the KGs exchange the $EM$s to encode questions and answer the questions from each other. Quality is evaluated by comparing the scores. Performances of the $QM$ and the $AM$ reveal the quality at ability level. There are only two models (i.e., $TM$ and $EM$) and two kinds of information (i.e., questions and answers) exchanged between KGs. Quality evaluation under such incomplete information effectively avoids exposure of internal information.
\begin{algorithm}[H]
	\caption{QEII}
	\label{alg:QEII}
	\textbf{Input}: KG $\alpha$ and $\beta$ \\
	\textbf{Output}: Evaluation results
	\begin{algorithmic}[1]
		\FOR {$triples\in \alpha\;or\;\beta$}
		\STATE Incrementally train $TM$;
		\ENDFOR
		\STATE $Q^{\alpha}_A,Q^{\alpha}_B\leftarrow Sample(\alpha)$;
		\FOR {$qa \in Q^{\alpha}_A$}
		\STATE Jointly train $EM^{\alpha}$ and $AM^{\alpha}$;
		\ENDFOR
		\FOR {$qa \in Q^{\alpha}_B$}
		\STATE Adversarially train $QM^{\alpha}$ and $AM^{\alpha}$;
		\ENDFOR
		\STATE Obtain question set $Q_{final}^{\alpha}$ finally generated by $QM^{\alpha}$;
		\STATE Similar to 4-11, train $EM^{\beta}$, $QM^{\beta}$ and $AM^{\beta}$, and obtain $Q_{final}^{\beta}$ using $\beta$;
		\STATE Exchange $EM^{\alpha}$ and $EM^{\beta}$;
		\STATE $\textbf{FSG}^{\alpha}\leftarrow EM^{\beta}(Q_{final}^{\alpha})$;
		\STATE $\textbf{FSG}^{\beta}\leftarrow EM^{\alpha}(Q_{final}^{\beta})$;
		\STATE Exchange $\textbf{FSG}^{\alpha}$ and $\textbf{FSG}^{\beta}$;
		\STATE $S^{\alpha}\leftarrow AM^{\alpha}(\textbf{FSG}^{\beta},\alpha)$;
		\STATE $S^{\beta}\leftarrow AM^{\beta}(\textbf{FSG}^{\alpha},\beta)$;
		\IF {$S^{\alpha} > S^{\beta}$}
		\STATE $\alpha$ is better;
		\ELSIF {$S^{\alpha} > S^{\beta}$}
		\STATE $\beta$ is better;
		\ELSE \STATE $\alpha$ and $\beta$ are of equal quality;
		\ENDIF
	\end{algorithmic}
\end{algorithm}

\section{Experiments and Discussions}\label{sec:four}
In this section, the proposed QEII is employed to conduct quality evaluation experiments on various KGs. Meanwhile, several baselines are also used to evaluate the KGs and are compared with ours. To ensure that the evaluation experiments are reasonable, question difficulty features used in the $QM$ and convergence of the trained $AM$ are analyzed in advance. Besides, common knowledge of the KGs and relationships between their quality and related statistics are also analyzed following the evaluation experiments.

\subsection{Datasets}
Experiments are conducted on 4 datasets: Harry Potter (H.P.$\alpha$ and H.P.$\beta$), Pokemon (P.K.M.$\alpha$ and P.K.M.$\beta$), Honglou (H.L.M.$\alpha$ and H.L.M.$\beta$), and Sanguo (S.G.$\alpha$ and S.G.$\beta$). Each dataset is a pair of KGs that belong to the same domain. All KGs are collected from OpenKG.CN\footnote{\url{http://www.openkg.cn/}}. Harry Potter, Honglou, and Sanguo mainly describe relationships between persons, while Pokemon primarily focuses on the dynamic between humans and elves. Statistics and construction date of these KGs are shown in Table \ref{tab:statistics}.
\begin{table}[!t]
	\caption{Statistics and construction date of datasets.}
	\label{tab:statistics}
	\centering
	\begin{tabular}{llcccc}
		\toprule
		\multicolumn{2}{c}{Datasets} & Entities & Relations & Triples & \makecell[c]{Construction \\ date} \\
		\midrule
		\multirow{2}{*}{\makecell[c]{Harry \\ Potter}} & H.P.$\alpha$ & 648 & 144 & 1738 & 2021.01.28 \\
		& H.P.$\beta$ & 814 & 200 & 2705 & 2021.01.26 \\
		\midrule
		\multirow{2}{*}{Pokemon} & P.K.M.$\alpha$ & 3708 & 18 & 75383 & 2021.01.28 \\
		& P.K.M.$\beta$ & 2589 & 7 & 15091 & 2021.01.25 \\
		\midrule
		\multirow{2}{*}{Honglou} & H.L.M.$\alpha$ & 388 & 47 & 380 & 2018.09.15 \\
		& H.L.M.$\beta$ & 284 & 44 & 7527 & 2021.12.10 \\
		\midrule
		\multirow{2}{*}{Sanguo} & S.G.$\alpha$ & 123 & 29 & 153 & 2018.09.15 \\
		& S.G.$\beta$ & 197 & 20 & 361 & 2021.11.30 \\
		\bottomrule
	\end{tabular}
\end{table}

\subsection{Experimental Settings}
Two question sets are sampled for model training. One is used to jointly train the $EM$ and the $AM$, and consists of 1000 questions. Of these, 80\% are used as a training set, 10\% as a validation set, and 10\% as a test set. The other is used to adversarially train the $QM$ and the $AM$, and also consists of 1000 questions. In the rule-based method, the $QM$ tunes the difficulties of these questions directly. In the Naive Bayes-based and retrieval-based method, this set is extended to a question base with 80000 questions, from which the $QM$ selects 1000 questions for each round of competitions with the $AM$. The target range of accuracy is set as $\eta=\left[0.5,0.52\right]$.

\subsection{Baselines}
We choose 4 evaluation methods as baselines, from which a total of 9 metrics are derived. They can be categorized into \textit{shallow} and \textit{deep} evaluations based on their mechanisms. The shallow ones are as follows,
\begin{itemize}
	\item[(1)] Downloads: it implies users' preferences while selecting KGs and reflects the KG quality.
\end{itemize}
Scale of a KG is also usually used to evaluate the quality. It can be measured by some statistics such as,
\begin{itemize}
	\item[(2)] EN (Entity Number): the number of entities.
	\item[(3)] RN (Relation Number): the number of relations.
	\item[(4)] TN (Triple Number): the number of triples.
\end{itemize}
KE (Knowledge Embedding)~\cite{pujara2017sparsity} proposed 4 metrics based on entropy and density,
\begin{itemize}
	\item[(5)] EE (Entity Entropy): the entity entropy is used to measure the diversity of entities in a KG,
	\begin{equation}
	P\left(e\right)=\frac{|t.s=e|+|t.o=e|}{|T|},
	\end{equation}
	\begin{equation}
	EE=\sum_{e\in E}{-P\left(e\right)\log{P(e)}},
	\end{equation}
	where, $t$ is a triple $\left(s,p,o\right)$, $e$ is an entity in the entity set $E$, $T$ is a triple set, and $|\cdot|$ denotes the size of a set.
	\item[(6)] RE (Relation Entropy): the relation entropy is used for an evaluation of relation diversity,
	\begin{equation}
	P\left(r\right)=\frac{|t.p=r|}{|T|},
	\end{equation}
	\begin{equation}
	RE=\sum_{r\in R}{-P\left(r\right)\log{P(r)}},
	\end{equation}
	where, $r$ is a relation in the relation set $R$.
	\item[(7)] ED (Entity Density): the entity density is used to measure the sparsity of entities in a KG,
	\begin{equation}
	ED=\frac{2|T|}{|E|}.
	\end{equation}
	\item[(8)] RD (Relation Density): the relation density is used for an evaluation of relation sparsity,
	\begin{equation}
	RD=\frac{|T|}{|R|}.
	\end{equation}
\end{itemize}

Shallow metrics evaluate KGs according to some shallow information such as the numbers of entities, relations, and triples. In contrast, deep methods always mine deep information from KGs for evaluation. We choose a deep baseline as follows,
\begin{itemize}
	\item[(9)] TTMF (Triple Trustworthiness Measurement)~\cite{jia2019triple}: it used neural networks to mine information from three aspects: i. is there a relation between two entities; ii. is a given relation the relation between these two entities; iii. can the related triples help predict that this triple is trustworthy. TTMF is used to evaluate whether a given triple is trustworthy. Quality of the whole KG is evaluated by the accuracy of trustworthy triples,
	\begin{equation}
	AccT=\frac{|CT|}{|T|},
	\end{equation}
	where $CT$ denotes trustworthy triples evaluated by TTMF and $T$ denotes all triples in the KG.
\end{itemize}

\subsection{Evaluation Metric}
Quality of a KG is reflected in performances of the $QM$ and the $AM$ trained on it. The question set generated by the $QM$ can be considered an examination paper, and a score is obtained after the $AM$ answers it. This score measures the performances of both models. The total score of a paper is typically 100. Suppose that each question is weighted equally, then percentage score is calculated as,
\begin{equation}
S=\frac{N_{Q^+}}{N_Q}\times 100,
\end{equation}
where $N_{Q^+}$ denotes the number of correctly answered questions and $N_Q$ is the total number of questions.

\subsection{Analysis for Model Training}
\subsubsection{Convergence Analysis for AM}
\begin{figure}[!t]
	\centering
	\includegraphics[width=0.99\linewidth]{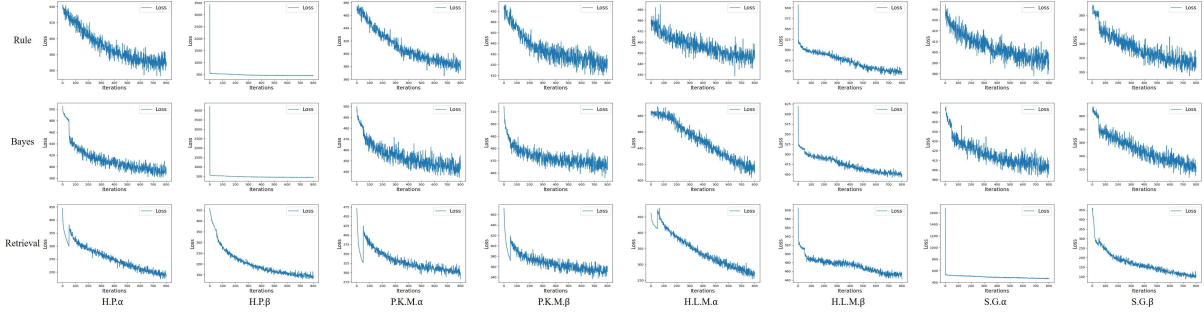}
	\caption{Training loss of AMs in the last round of adversarial processes.}
	\label{fig:convergence}
\end{figure}

In the first subgame, the $AM$ learns to answer the questions generated by the $QM$. Training is halted when accuracy of the $AM$ is within the target range or the number of competition rounds reaches the limit. To confirm $AM$ convergence, the training loss of the $AM$ in the last round of competitions is recorded, as displayed in Figure \ref{fig:convergence}. The results demonstrate that all $AM$s have converged.

\subsubsection{Analysis for Question Difficulty Features}
\begin{figure}[!t]
	\centering
	\includegraphics[width=0.6\linewidth]{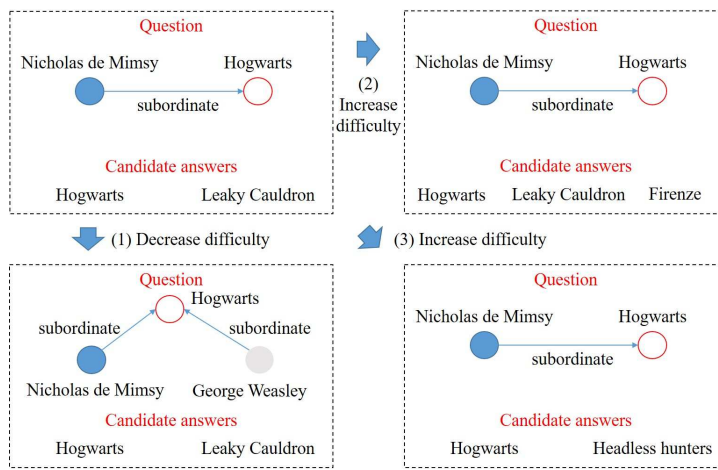}
	\caption{Examples of question difficulty features.}
	\label{fig:dexample}
\end{figure}

Questions represented in natural language may have different difficulties based on how they are described. Differently, a defect subgraph cannot be endowed with different difficulties through different structures since it describes a fact. Therefore, three difficulty features are proposed as follows, whose examples are shown in Figure \ref{fig:dexample}.
\begin{itemize}
	\item[(1)] The amount of information directly related to the answer in the question. The greater it is, the lower the question difficulty.
	\item[(2)] The number of candidate answers. The greater it is, the higher the question difficulty.
	\item[(3)] The relevance of candidate answers to the question. The greater it is, the smaller the distinction between correct and wrong answers, and the higher the question difficulty.
\end{itemize}

\begin{figure}[!t]
	\centering
	\includegraphics[width=0.6\linewidth]{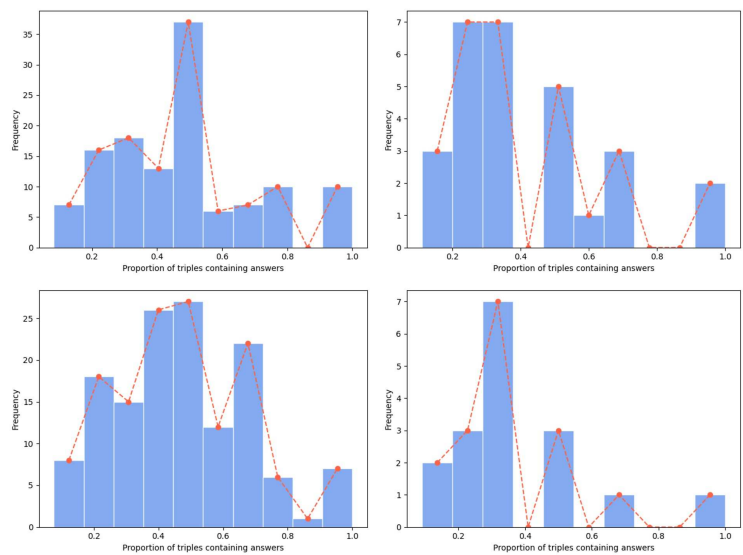}
	\caption{Proportion of triples containing answers in correctly (left) and wrongly (right) answered questions on H.P.$\alpha$ (top) and H.P.$\beta$ (bottom).}
	\label{fig:feature1}
\end{figure}

To assess and confirm the validity of the aforementioned difficulty features, 200 questions are selected randomly from the question set utilized for adversarial training on Harry Potter, and the $AM$ is employed to answer these questions.
\begin{itemize}
	\item[(1)] The questions are divided into correctly and wrongly answered parts after being answered. Figure \ref{fig:feature1} shows the proportion of triples containing answers in the subgraphs. These results demonstrate that the proportion is greater than 0.4 in most correctly answered questions but is less than 0.4 in most wrongly answered questions. The triples containing answers denote information directly related to the answers. The question difficulty is related to the amount of this information.
	\item[(2)] Some entities are selected randomly as wrong candidates to generate choice questions with 2, 3, 4, and 5 candidate answers. The $AM$ answers these questions respectively and the results are shown in Table \ref{tab:feature2}. As the number of candidate answers increases, there is a decrease in the number of correctly answered questions and an increase in the number of wrongly answered questions. This suggests that the question difficulty is related to the number of candidate answers.
	\item[(3)] An entity is selected randomly from i. the KG, ii. the question, and iii. entities directly related to the question (i.e., neighbors of the defect subgraph) as a wrong candidate, and the $AM$ is then used to answer these choice questions with 2 candidate answers. The results are shown in Table \ref{tab:feature3}, which demonstrate that the number of correctly answered questions decreases significantly when the wrong candidate comes from ii or iii. The question difficulty is related to the relevance of candidate answers to the question.
\end{itemize}
\begin{table}[!t]
	\caption{Results of answering questions with different numbers of candidates.}
	\label{tab:feature2}
	\centering
	\begin{tabular}{lcccc}
		\toprule
		\multirow{2}{*}{\makecell[c]{Number of \\ candidates}} & \multicolumn{2}{c}{H.P.$\alpha$} & \multicolumn{2}{c}{H.P.$\beta$} \\
		\cmidrule(lr){2-3} \cmidrule(lr){4-5}
		& correct & wrong & correct & wrong \\
		\midrule
		2 & 120 & 80 & 151 & 49 \\
		3 & 104 & 96 & 120 & 80 \\
		4 & 79 & 121 & 95 & 105 \\
		5 & 75 & 125 & 92 & 108 \\
		\bottomrule
	\end{tabular}
\end{table}
\begin{table}[!t]
	\caption{Results of answering questions with candidates from different sources.}
	\label{tab:feature3}
	\centering
	\begin{tabular}{lcccc}
		\toprule
		\multirow{2}{*}{\makecell[c]{Source of \\ candidates}} & \multicolumn{2}{c}{H.P.$\alpha$} & \multicolumn{2}{c}{H.P.$\beta$} \\
		\cmidrule(lr){2-3} \cmidrule(lr){4-5}
		& correct & wrong & correct & wrong \\
		\midrule
		i & 120 & 80 & 151 & 49 \\
		ii & 88 & 112 & 97 & 103 \\
		iii & 77 & 123 & 105 & 95 \\
		\bottomrule
	\end{tabular}
\end{table}

\subsection{Experimental Results}
In the evaluation experiments, each KG not only answers questions from the other KG but also answers those from itself. To avoid occasionality, we sample 9 additional question sets whose difficulties are similar to the questions generated by the $QM$. The $AM$ answers questions from 10 sets respectively and the average scores are shown in Table \ref{tab:QEII}. The numbers in bold represent greater scores after KGs answer each other's questions. The numbers with underlines represent greater scores after KGs answer their own questions. The numbers with * represent greater scores after a KG answers questions from the other KG and itself.
\begin{table}[!t]
	\caption{Evaluation results of QEII.}
	\label{tab:QEII}
	\centering
	\begin{tabular}{lcccccc}
		\toprule
		\multirow{2}{*}{KGs} & \multicolumn{3}{c}{Answer own's} & \multicolumn{3}{c}{Answer the other's} \\
		\cmidrule(lr){2-4}\cmidrule(lr){5-7}
		& Rule & Bayes & Retrieval & Rule & Bayes & Retrieval \\
		\midrule
		H.P.$\alpha$	&	\underline{44.7}*	&	\underline{39.2}*	&	31.8*	&	31.8	&	33.1	&	\textbf{29.0}	\\
		H.P.$\beta$	&	42.4*	&	37.8*	&	\underline{32.0}*	&	\textbf{39.0}	&	\textbf{33.7}	&	22.9	\\
		\midrule
		P.K.M.$\alpha$	&	40.0*	&	\underline{44.7}*	&	\underline{36.5}*	&	35.4	&	\textbf{38.7}	&	\textbf{32.3}	\\
		P.K.M.$\beta$	&	\underline{41.9}*	&	37.3*	&	35.8*	&	\textbf{36.6}	&	34.0	&	29.9	\\
		\midrule
		H.L.M.$\alpha$	&	44.0*	&	35.9*	&	31.6	&	\textbf{42.3}	&	33.4	&	\textbf{34.8}*	\\
		H.L.M.$\beta$	&	\underline{54.9}*	&	\underline{46.7}*	&	\underline{46.9}*	&	35.2	&	\textbf{38.2}	&	30.3	\\
		\midrule
		S.G.$\alpha$	&	54.9*	&	54.7*	&	\underline{54.9}*	&	39.7	&	40.5	&	30.5	\\
		S.G.$\beta$	&	\underline{57.8}*	&	\underline{60.6}*	&	50.8*	&	\textbf{49.7}	&	\textbf{50.5}	&	\textbf{48.3}	\\
		\bottomrule
	\end{tabular}
\end{table}

The scores of KGs to answer their own questions are typically greater because the $AM$ and the questions are from the same KGs. These scores are not appropriate for quality evaluation. They measure the performance of the $AM$ but not the $QM$, which conflicts with the intended purpose of our evaluation. The scores of KGs to answer each other's questions reflect the answering ability of the $AM$ and the question ability of the $QM$ from the other one. Therefore, these scores are finally used to evaluate the quality of KGs. Experimental results demonstrate that the quality of H.P.$\beta$ is better than that of H.P.$\alpha$, P.K.M.$\alpha$ is better than P.K.M.$\beta$, H.L.M.$\alpha$ is better than H.L.M.$\beta$, and S.G.$\beta$ is better than S.G.$\alpha$.

The shallow and deep baselines are also used to evaluate the same KGs, whose results are shown in Table \ref{tab:baseline}.
\begin{table}[!t]
	\caption{Evaluation results of baselines.}
	\label{tab:baseline}
	\centering
	\resizebox{1.0\linewidth}{!}{
		\begin{tabular}{llcccccccc}
			\toprule
			\multirow{2}{*}{Metric types} & \multirow{2}{*}{Metrics} & \multicolumn{2}{c}{Harry Potter} & \multicolumn{2}{c}{Pokemon} & \multicolumn{2}{c}{Honglou} & \multicolumn{2}{c}{Sanguo} \\
			\cmidrule(lr){3-4}\cmidrule(lr){5-6}\cmidrule(lr){7-8}\cmidrule(lr){9-10}
			& & H.P.$\alpha$ & H.P.$\beta$ & P.K.M.$\alpha$ & P.K.M.$\beta$ & H.L.M.$\alpha$ & H.L.M.$\beta$ & S.G.$\alpha$ & S.G.$\beta$ \\
			\midrule
			\multirow{8}{*}{Shallow} & Downloads & \textbf{695} & 141 & \textbf{193} & 83 & \textbf{2455} & 493 & \textbf{2455} & 101 \\
			& EN & 648 & \textbf{814} & \textbf{3708} & 2589 & \textbf{388} & 284 & 123 & \textbf{197} \\
			& RN & 144 & \textbf{200} & \textbf{18} & 7 & \textbf{47} & 44 & \textbf{29} & 20 \\
			& TN & 1738 & \textbf{2705} & \textbf{73781} & 15088 & 371 & \textbf{1702} & 146 & \textbf{359} \\
			& EE & 9.7934 & \textbf{10.4382} & \textbf{13.3548} & 13.0269 & \textbf{9.7738} & 8.1511 & 7.2281 & \textbf{7.8407} \\
			& RE & 3.3954 & \textbf{3.7570} & \textbf{1.7803} & 1.4370 & 2.6110 & \textbf{2.6220} & \textbf{2.4087} & 1.7071 \\
			& ED & \textbf{12.0694} & 6.5553 & \textbf{39.7956} & 11.6555 & 1.9124 & \textbf{11.9859} & 2.3740 & \textbf{3.6447} \\
			& RD & 5.3642 & \textbf{13.3400} & \textbf{4098.9444} & 2155.4286 & 7.8936 & \textbf{38.6818} & 5.0345 & \textbf{17.9500} \\
			\midrule
			Deep & AccT & 0.9813 & \textbf{0.9841} & 0.9486 & \textbf{0.9843} & \textbf{0.9899} & 0.9178 & 0.9701 & \textbf{0.9792} \\
			\bottomrule
		\end{tabular}
	}
\end{table}
Evaluation results of the shallow baselines demonstrate that the quality of H.P.$\beta$ is better than that of H.P.$\alpha$, P.K.M.$\alpha$ is better than P.K.M.$\beta$, H.L.M.$\alpha$ is similar with H.L.M.$\beta$, and S.G.$\beta$ is better than S.G.$\alpha$. Results of the deep baselines demonstrate that the quality of H.P.$\beta$ is better than that of H.P.$\alpha$, P.K.M.$\beta$ is better than P.K.M.$\alpha$, H.L.M.$\alpha$ is better than H.L.M.$\beta$, and S.G.$\beta$ is better than S.G.$\alpha$. The deep baselines are more convincing than the shallow ones because they make full use of deep information mined from KGs for quality evaluation. Their evaluation results are the same as ours except Pokemon. It is worth noting that all the shallow baselines give the same evaluation results on Pokemon and show that P.K.M.$\alpha$ is better than P.K.M.$\beta$, which is the same as ours.

However, the baselines have to rely on raw data in KGs. For example, entropy metrics, density metrics, and accuracy of the trustworthy triples all require detailed information about every triple, which leads to information exposure and privacy disclosure. In contrast, the QEII is able to achieve almost consistent evaluation results with the deep baselines under incomplete information. What's more, the evaluation of the QEII is at ability level, which is fundamentally different from the baselines. The QEII is much more effective and superior.

\subsection{Analysis and Discussion}
\subsubsection{Analysis for Common Knowledge}
There exist many differences in knowledge between a pair of KGs although they both belong to the same domain. While answering each other's questions, the $AM$ may fail to correctly answer some questions due to the lack of relevant knowledge in its trained KG. To further explore the answering ability of the $AM$, we construct a common KG by extracting common knowledge or identical triples from both KGs, and sample questions from it. The trained $AM$ is used to answer these questions. To minimize randomness, the sampling process is repeated 10 times, with each time comprising 1000 questions. Table \ref{tab:analysis1} presents the average scores obtained. While these results are useful in evaluating quality, the evaluation cannot rely solely on them. An $AM$ trained using a particular KG may possess better generalization ability, allowing it to answer questions that go beyond the knowledge of the KG. This ability can also reflect the KG quality at ability level.
\begin{table}[!t]
	\caption{Results of answering questions from common knowledge.}
	\label{tab:analysis1}
	\centering
	\begin{tabular}{lccc}
		\toprule
		KGs & Rule & Bayes & Retrieval \\
		\midrule
		H.P.$\alpha$	&	32.4	&	32.8	&	36.8	\\
		H.P.$\beta$	&	\textbf{37.4}	&	\textbf{36.9}	&	\textbf{37.9}	\\
		\midrule
		P.K.M.$\alpha$	&	\textbf{35.4}	&	\textbf{34.1}	&	\textbf{35.0}	\\
		P.K.M.$\beta$	&	33.5	&	32.5	&	32.2	\\
		\midrule
		H.L.M.$\alpha$	&	36.7	&	27.9	&	28.5	\\
		H.L.M.$\beta$	&	\textbf{39.8}	&	\textbf{40.2}	&	\textbf{40.8}	\\
		\midrule
		S.G.$\alpha$	&	40.2	&	44.1	&	45.6	\\
		S.G.$\beta$	&	\textbf{54.6}	&	\textbf{51.1}	&	\textbf{58.8}	\\
		\bottomrule
	\end{tabular}
\end{table}

\subsubsection{Analysis for Relationships between KG Quality and Statistics}
To explore relationships between KG quality and statistics, we remove some triples from the KG with better quality and evaluate this KG again. Harry Potter is used for experiments and the rule-based method is used to tune question difficulty. 300, 600, 900, 1200, and 1500 triples are removed from H.P.$\beta$ respectively. There are two types of triples in H.P.$\beta$: unique triples and common triples with H.P.$\alpha$. The removal of these triples is based on the ratio of their amount, which is about 4:1. It is worth noting that the number of triples in H.P.$\beta$ is less than that in H.P.$\alpha$ while removing 1200 and 1500 triples. To eliminate any chance of occasional biases, triple removal and quality evaluation are repeated 10 times, whose average results are shown in Table \ref{tab:analysis2}.
\begin{table}[!t]
	\caption{Evaluation results after removing different numbers of triples.}
	\label{tab:analysis2}
	\centering
	\begin{tabular}{lcccc}
		\toprule
		\multirow{2}{*}{\makecell[l]{Removed \\ triples}} & \multicolumn{2}{c}{Q-H.P.$\alpha$ A-H.P.$\beta$} & \multicolumn{2}{c}{Q-H.P.$\beta$ A-H.P.$\alpha$} \\
		\cmidrule(lr){2-3} \cmidrule(lr){4-5}
		&	avg.	&	std.	&	avg.	&	std.	\\
		\midrule
		300	&	\textbf{40.2}	&	21.6	&	37.1	&	28.4	\\
		600	&	\textbf{50.4}	&	27.1	&	47.4	&	30.1	\\
		900	&	\textbf{33.5}	&	23.6	&	32.6	&	23.7	\\
		1200	&	\textbf{32.5}	&	12.2	&	31.1	&	8.5	\\
		1500	&	26.6	&	25.3	&	\textbf{28.5}	&	10.6	\\
		\bottomrule
	\end{tabular}
\end{table}
The results demonstrate that while removing 1200 triples, H.P.$\beta$ is better than H.P.$\alpha$ although the number of triples in H.P.$\beta$ is less than that in H.P.$\alpha$. This suggests that an assessment of KG quality based solely on statistics may be insufficient, and that completeness and integrity of knowledge should also be considered as crucial factors.

\section{Conclusion}\label{sec:five}
In this paper, we propose a knowledge graph quality evaluation framework under incomplete information (QEII). Different from existing methods, the QEII focuses on the quality at ability level. It transforms the quality evaluation task into an adversarial Q\&A game between KGs. Each KG uses its own knowledge to train a $QM$ and an $AM$ in a GAN manner. In the mutual Q\&A, KGs exchange questions for answering and exchange the answers to give a score. The quality can be evaluated by comparing the scores. Performances of the models reflect ability of the KGs. Questions, answers, $TM$, and $EM$ are the only information exchanged between KGs, without exposing any internal information. In the evaluation experiments, we compare the QEII with several shallow and deep baselines. The experimental results demonstrate that the evaluation results of the QEII are almost consistent with the deep baselines, but the QEII implements the evaluation under incomplete information. In addition, we also make two analyses: analysis for common knowledge of KGs indicates that this knowledge helps evaluate the quality; analysis for relationships between KG quality and statistics indicates that the quality evaluation based on statistics alone is not sufficient.

In future work, we will investigate the Q\&A interaction between KGs in different representations and enhance the adversarial training between $QM$ and $AM$ to achieve greater flexibility in the evaluation process.

\bibliographystyle{unsrt}  
\bibliography{references}

\end{document}